\documentclass{article}
\usepackage{arXiv_message}
\usepackage[english]{babel}
\usepackage[utf8]{inputenc} 
\usepackage[T1]{fontenc}    
\usepackage{hyperref}       
\usepackage{url}            
\usepackage{booktabs}       
\usepackage{amsfonts}       
\usepackage{nicefrac}       
\usepackage{microtype}      
\usepackage[colorinlistoftodos]{todonotes}
\usepackage{wrapfig}
\usepackage{soul}
\usepackage{array}
\usepackage{amsmath}
\DeclareMathOperator*{\argmin}{argmin}
\usepackage{multirow}

\title{Three scenarios for continual learning}
\author{Gido M. van de Ven$^{1,2}$ \ \& Andreas S. Tolias$^{1,3}$ \\
$^1$ Center for Neuroscience and Artificial Intelligence, Baylor College of Medicine, Houston \\
$^2$ Computational and Biological Learning Lab, University of Cambridge, Cambridge \\
$^3$ Department of Electrical and Computer Engineering, Rice University, Houston \\
\texttt{\{ven,astolias\}@bcm.edu} \\
}
\begin{document}

\maketitle

\begin{abstract}
Standard artificial neural networks suffer from the well-known issue of catastrophic forgetting, making continual or lifelong learning difficult for machine learning. In recent years, numerous methods have been proposed for continual learning, but due to differences in evaluation protocols it is difficult to directly compare their performance. To enable more structured comparisons, we describe three continual learning scenarios based on whether at test time task identity is provided and---in case it is not---whether it must be inferred. Any sequence of well-defined tasks can be performed according to each scenario. Using the split and permuted MNIST task protocols, for each scenario we carry out an extensive comparison of recently proposed continual learning methods. We demonstrate substantial differences between the three scenarios in terms of difficulty and in terms of how efficient different methods are. In particular, when task identity must be inferred (i.e., class incremental learning), we find that regularization-based approaches (e.g., elastic weight consolidation) fail and that replaying representations of previous experiences seems required for solving this scenario.
\end{abstract}

\section{Introduction}
\label{sec:introduction}

Current state-of-the-art deep neural networks can be trained to impressive performance on a wide variety of individual tasks. Learning multiple tasks in sequence, however, remains a substantial challenge for deep learning. When trained on a new task, standard neural networks forget most of the information related to previously learned tasks, a phenomenon referred to as ``catastrophic forgetting''.

In recent years, numerous methods for alleviating catastrophic forgetting have been proposed. However, due to the wide variety of experimental protocols used to evaluate them, many of these methods claim ``state-of-the-art'' performance [e.g., \citealp{kirkpatrick2017overcoming,rebuffi2017icarl,nguyen2017variational,masse2018alleviating,kemker2018fearnet,wu2018incremental}]. To obscure things further, some methods shown to perform well in some experimental settings are reported to dramatically fail in others: compare the performance of elastic weight consolidation in \citet{kirkpatrick2017overcoming} and \citet{zenke2017improved} with that in \citet{kemker2017measuring} and \citet{kamra2017deep}.

To enable more structured comparisons of methods for reducing catastrophic forgetting, this report describes three distinct continual learning scenarios of increasing difficulty. These scenarios are distinguished by whether at test time task identity is provided and, if it is not, whether task identity must be inferred. We show that there are substantial differences between these three scenarios in terms of their difficulty and in terms of how effective different continual learning methods are on them. Moreover, using the split and permuted MNIST task protocols, we illustrate that any continual learning problem consisting of a series of clearly separated tasks can be performed according to all three scenarios.

As a second contribution, for each of the three scenarios we then provide an extensive comparison of recently proposed methods. These experiments reveal that even for experimental protocols involving the relatively simple classification of MNIST-digits, regularization-based approaches (e.g., elastic weight consolidation) completely fail when task identity needs to be inferred. We find that currently only replay-based approaches have the potential to perform well on all three scenarios. Well-documented and easy-to-adapt code for all compared methods is made available: \url{https://github.com/GMvandeVen/continual-learning}.

\section{Three Continual Learning Scenarios}
\label{sec:scenarios}

We focus on the continual learning problem in which a single neural network model needs to sequentially learn a series of tasks. During training, only data from the current task is available and the tasks are assumed to be clearly separated. This problem has been actively studied in recent years and many methods for alleviating catastrophic forgetting have been proposed. However, because of differences in the experimental protocols used for their evaluation, comparing methods' performances can be difficult. In particular, one difference between experimental protocols we found to be very influential for the level of difficulty is whether at test time information about the task identity is available and---if it is not---whether the model is also required to explicitly identify the identity of the task it has to solve. Importantly, this means that even when studies use exactly the same sequence of tasks to be learned (i.e., the same task protocol), results are not necessarily comparable. In the hope to standardize evaluation and to enable more meaningful comparisons across papers, we describe three distinct scenarios for continual learning of increasing difficulty (Table \ref{tab:scenarios}). This categorization scheme was first introduced in a previous paper by us \citep{van2018generative} and has since been adopted by several other studies \citep{hsu2018re,zeno2019task,lee2019incremental}; the purpose of the current report is to provide a more in-depth treatment.

\begin{table}[h]
  \caption{\label{tab:scenarios}Overview of the three continual learning scenarios.}
  \begin{center}
  \begin{tabular}{ll}
    \toprule
    \textbf{\emph{Scenario}} & \emph{\textbf{Required at test time}} \\
    \midrule \midrule
    \textbf{Task-IL} & Solve tasks so far, task-ID provided \\
    \midrule
    \textbf{Domain-IL} & Solve tasks so far, task-ID not provided\\
    \midrule
    \textbf{Class-IL} & Solve tasks so far \emph{and} infer task-ID \\
    \bottomrule
  \end{tabular}
  \end{center}
\end{table}

In the first scenario, models are always informed about which task needs to be performed. This is the easiest continual learning scenario, and we refer to it as \textbf{task-incremental learning (Task-IL)}. Since task identity is always provided, in this scenario it is possible to train models with task-specific components. A typical network architecture used in this scenario has a ``multi-headed'' output layer, meaning that each task has its own output units but the rest of the network is (potentially) shared between tasks.

In the second scenario, which we refer to as \textbf{domain-incremental learning (Domain-IL)}, task identity is not available at test time. Models however only need to solve the task at hand; they are not required to infer which task it is. Typical examples of this scenario are protocols whereby the structure of the tasks is always the same, but the input-distribution is changing. A relevant real-world example is an agent who needs to learn to survive in different environments, without the need to explicitly identify the environment it is confronted with.

Finally, in the third scenario, models must be able to both solve each task seen so far \emph{and} infer which task they are presented with. We refer to this scenario as \textbf{class-incremental learning (Class-IL)}, as it includes the common real-world problem of incrementally learning new classes of objects.

\subsection{Comparison with Single-Headed vs Multi-Headed Categorization Scheme}
In another recent attempt to structure the continual learning literature, a distinction is highlighted between methods being evaluated using a ``multi-headed'' or a ``single-headed'' layout \citep{farquhar2018towards,chaudhry2018riemannian}. This distinction relates to the scenarios we describe here in the sense that a multi-headed layout requires task identity to be known, while a single-headed layout does not. Our proposed categorization however differs in two important ways.

Firstly, the multi-headed vs single-headed distinction is tied to the architectural layout of a network's output layer, while our scenarios more generally reflect the conditions under which a model is evaluated. Although in the continual learning literature a multi-headed layout (i.e., using a separate output layer for each task) is the most common way to use task identity information, it is not the only way. Similarly, a single-headed layout (i.e., using the same output-layer for every task) might by itself not require task identity to be known, it is still possible for the model to use task identity in other ways (e.g., in its hidden layers, as in \citep{masse2018alleviating}).

\begin{figure}[t]
  \begin{center}
  \centerline{\includegraphics[width=0.8\columnwidth]{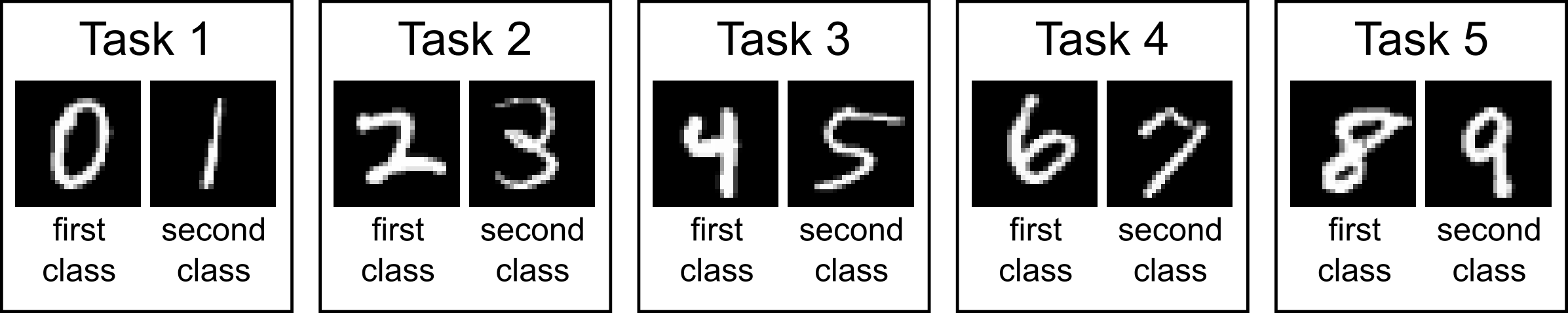}}
  \caption{\label{fig:exampleSplit}Schematic of split MNIST task protocol.}
  \end{center}
  \vskip -0.1in
\end{figure}

\begin{table}[t]
  \caption{\label{tab:scenariosSplit}Split MNIST according to each scenario.}
  \begin{center}
  \begin{tabular}{ll}
    \toprule
    \multirow{2}{*}{\bf Task-IL}   & With task given, is it the 1$^{\text{st}}$ or 2$^{\text{nd}}$ class? \\
    \textbf{}    & (e.g., $0$ or $1$) \\
    \midrule
    \multirow{2}{*}{\bf Domain-IL}   & With task unknown, is it a 1$^{\text{st}}$ or 2$^{\text{nd}}$ class?   \\
    & (e.g., in $[0, 2, 4, 6, 8]$ or in $[1, 3, 5, 7, 9]$) \\
    \midrule
    \multirow{2}{*}{\bf Class-IL}    & With task unknown, which digit is it? \\
    & (i.e., choice from $0$ to $9$) \\
    \bottomrule
  \end{tabular}
  \end{center}
  \vskip -0.05in
\end{table}

Secondly, our categorization scheme extends upon the multi-headed vs single-headed split by recognizing that when task identity is not provided, there is a further distinction depending on whether the network is explicitly required to infer task identity. Importantly, we will show that the two scenarios resulting from this additional split substantially differ in difficulty (see section~\ref{sec:results}).

\subsection{Example Task Protocols}
\label{sec:exampletasks}
To demonstrate the difference between the three continual learning scenarios, and to illustrate that any task protocol can be performed according to each scenario, we will perform two different task protocols for all three scenarios.

The first task protocol is sequentially learning to classify MNIST-digits (`split MNIST' \citep{zenke2017improved}; Figure~\ref{fig:exampleSplit}). In the recent literature, this task protocol has been performed under the Task-IL scenario (in which case it is sometimes referred to as `multi-headed split MNIST') and under the Class-IL scenario (in which case it is referred to as `single-headed split MNIST'), but it could also be performed under the Domain-IL scenario (Table~\ref{tab:scenariosSplit}).

\begin{figure}[t]
  \begin{center}
  \centerline{\includegraphics[width=0.8\columnwidth]{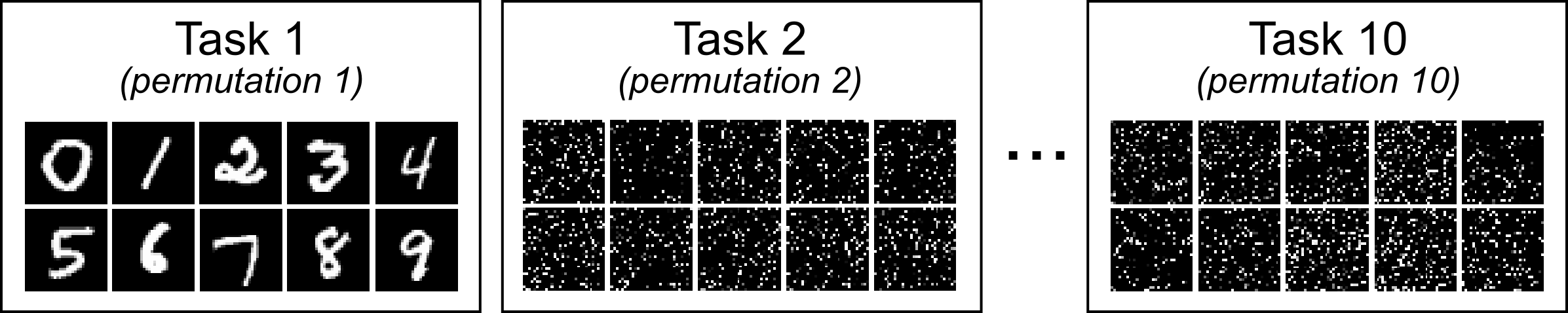}}
  \caption{\label{fig:examplePerm}Schematic of permuted MNIST task protocol.}
  \end{center}
  \vskip -0.1in
\end{figure}

\begin{table}[t]
  \caption{\label{tab:scenariosPerm}Permuted MNIST according to each scenario.}
  \begin{center}
  \begin{tabular}{ll}
    \toprule
    \textbf{Task-IL}     & Given permutation \emph{X}, which digit?\\
    \midrule
    \textbf{Domain-IL}   & With permutation unknown, which digit?   \\
    \midrule
    \textbf{Class-IL}    & Which digit \emph{and} which permutation? \\
    \bottomrule
  \end{tabular}
  \end{center}
  \vskip -0.05in
\end{table}

The second task protocol is `permuted MNIST' \citep{goodfellow2013empirical}, in which each task involves classifying all ten MNIST-digits but with a different permutation applied to the pixels for every new task (Figure~\ref{fig:examplePerm}). Although permuted MNIST is most naturally performed according to the Domain-IL scenario, it can be performed according to the other scenarios too (Table~\ref{tab:scenariosPerm}).

\subsection{Task Boundaries}
The scenarios described in this report assume that \emph{during training} there are clear and well-defined boundaries between the tasks to be learned. If there are no such boundaries between tasks---for example because transitions between tasks are gradual or continuous---the scenarios we describe here no longer apply, and the continual learning problem becomes less structured and potentially a lot harder. Among others, training with randomly-sampled minibatches and multiple passes over each task's training data are no longer possible. We refer to \cite{zeno2019task} for a recent insightful treatment of the paradigm without well-defined task-boundaries.


\section{Strategies for Continual Learning}
\label{sec:strategies}

\subsection{Task-specific Components}
A simple explanation for catastrophic forgetting is that after a neural network is trained on a new task, its parameters are optimized for the new task and no longer for the previous one(s). This suggests that not optimizing the entire network on each task could be one strategy for alleviating catastrophic forgetting. A straightforward way to do this is to explicitly define a different sub-network per task. Several recent papers use this strategy, with different approaches for selecting the parts of the network for each task. A simple approach is to randomly assign which nodes participate in each task (\emph{Context-dependent Gating} [XdG; \citealp{masse2018alleviating}]). Other approaches use evolutionary algorithms \citep{fernando2017pathnet} or gradient descent \citep{serra2018overcoming} to learn which units to employ for each task. By design, these approaches are limited to the Task-IL scenario, as task identity is required to select the correct task-specific components.

\subsection{Regularized Optimization}
When task identity information is not available at test time, an alternative strategy is to still preferentially train a different part of the network for each task, but to always use the entire network for execution. One way to do this is by differently regularizing the network's parameters during training on each new task, which is the approach of \emph{Elastic Weight Consolidation} [EWC; \citealp{kirkpatrick2017overcoming}] and \emph{Synaptic Intelligence} [SI; \citealp{zenke2017improved}]. Both methods estimate for all parameters of the network how important they are for the previously learned tasks and penalize future changes to them accordingly (i.e., learning is slowed down for parts of the network important for previous tasks).

\subsection{Modifying Training Data}
An alternative strategy for alleviating catastrophic forgetting is to complement the training data for each new task to be learned with ``pseudo-data'' representative of the previous tasks. This strategy is referred to as replay.

One option is to take the input data of the current task, label them using the model trained on the previous tasks, and use the resulting input-target pairs as pseudo-data. This is the approach of \emph{Learning without Forgetting} [LwF; \citealp{li2017learning}]. An important aspect of this method is that instead of labeling the replayed inputs as the most likely category according to the previous tasks' model (i.e., ``hard targets''), it pairs them with the predicted probabilities for \emph{all} target classes (i.e., ``soft targets''). The objective for the replayed data is to match the probabilities predicted by the model being trained to these target probabilities. The approach of matching predicted probabilities of one network to those of another network had previously been used to compress (or ``distill'') information from one (large) network to another (smaller) network \citep{hinton2015distilling}.

An alternative is to generate the input data to be replayed. For this, besides the main model for task performance (e.g., classification), a separate generative model is sequentially trained on all tasks to generate samples from their input data distributions. For the first application of this approach, which was called \emph{Deep Generative Replay} (DGR), the generated input samples were paired with ``hard targets'' provided by the main model \citep{shin2017continual}. We note that it is possible to combine DGR with distillation by replaying input samples from a generative model and pairing them with soft targets [see also \citealp{wu2018incremental,venkatesan2017strategy}]. We include this hybrid method in our comparison under the name DGR+distill.

A final option is to store data from previous tasks and replay those. Such ``exact replay'' has been used in various forms to successfully boost continual learning performance in classification settings [e.g., \citealp{rebuffi2017icarl,nguyen2017variational,kemker2018fearnet}]. A disadvantage of this approach is that it is not always possible, for example due to privacy concerns or memory constraints.

\subsection{Using Exemplars}
If it is possible to store data from previous tasks, another strategy for alleviating catastrophic forgetting is to use stored data as ``exemplars'' during execution. A recent method that successfully used this strategy is \emph{iCaRL} \citep{rebuffi2017icarl}. This method uses a neural network for feature extraction and performs classification based on a nearest-class-mean rule \citep{mensink2012metric} in that feature space, whereby the class means are calculated from the stored data. To protect the feature extractor network from becoming unsuitable for previously learned tasks, iCaRL also replays the stored data---as well as the current task inputs with a special form of distillation---during training of the feature extractor.

\section{Experimental Details}
\label{sec:expdetails}

In order to both explore the differences between the three continual learning scenarios and to comprehensively compare the performances of the above discussed approaches, we evaluated various recently proposed methods according to each scenario on both the split and permuted MNIST task protocols.

\subsection{Task Protocols}
For split MNIST, the original MNIST-dataset was split into five tasks, where each task was a two-way classification. The original 28x28 pixel grey-scale images were used without pre-processing. The standard training/test-split was used resulting in 60,000 training images ({\raise.17ex\hbox{$\scriptstyle\sim$}}6000 per digit) and 10,000 test images ({\raise.17ex\hbox{$\scriptstyle\sim$}}1000 per digit).

For permuted MNIST, a sequence of ten tasks was used. Each task is now a ten-way classification. To generate the permutated images, the original images were first zero-padded to 32x32 pixels. For each task, a random permutation was then generated and applied to these 1024 pixels. No other pre-processing was performed. Again the standard training/test-split was used.

\subsection{Methods}
\label{sec:methods}
For a fair comparison, the same neural network architecture was used for all methods. This was a multi-layer perceptron with 2 hidden layers of 400 (split MNIST) or 1000 (permuted MNIST) nodes each. ReLU non-linearities were used in all hidden layers. Except for iCaRL, the final layer was a softmax output layer. In the Task-IL scenario, all methods used a multi-headed output layer, meaning that each task had its own output units and always only the output units of the task under consideration---i.e., either the current task or the replayed task---were active. In the Domain-IL scenario, all methods were implemented with a single-headed output layer, meaning that each task used the same output units (with each unit corresponding to one class in every task). In the Class-IL scenario, each class had its own output unit and always all units of the classes seen so far were active (see also sections \ref{sec:classifcation} and \ref{sec:distillation} in the Appendix).

We compared the following methods:
\begin{description}
\item[- XdG:] Following \citet{masse2018alleviating}, for each task a random subset of $X\%$ of the units in each hidden layer was fully gated (i.e., their activations set to zero), with $X$ a hyperparameter whose value was set by a grid search (see section~\ref{sec:hyper} in the Appendix). As this method requires availability of task identity at test time, it can only be used in the Task-IL scenario.
\item[- EWC / Online EWC / SI:] For these methods a regularization term was added to the loss, with regularization strength controlled by a hyperparameter: $\mathcal{L}_{\text{total}} = \mathcal{L}_{\text{current}} + \lambda \mathcal{L}_{\text{regularization}}$. The value of this hyperparameter was again set by a grid search. The way the regularization terms of these methods are calculated differs (\cite{kirkpatrick2017overcoming,schwarz2018progress,zenke2017improved}; see section~\ref{sec:regularization} in the Appendix), but they all aim to penalize changes to parameters estimated to be important for previously learned tasks.
\item[- LwF / DGR / DGR+distill:] For these methods a loss-term for replayed data was added to the loss of the current task. In this case a hyperparameter could be avoided, as the loss for the current and replayed data was weighted according to how many tasks the model had been trained on so far: $\mathcal{L}_{\text{total}} = \frac{1}{N_{\text{tasks so far}}} \mathcal{L}_{\text{current}} + (1-\frac{1}{N_{\text{tasks so far}}}) \mathcal{L}_{\text{replay}}$.
\begin{itemize}
\item For LwF, images of the current task were replayed with soft targets provided by a copy of the model stored after finishing training on the previous task (\cite{li2017learning}; see also section~\ref{sec:distillation} in the Appendix).
\item For DGR, a separate generative model (see below) was trained to generate the images to be replayed. Following \citet{shin2017continual}, the replayed images were labeled with the most likely category predicted by a copy of the main model stored after training on the previous task (i.e., hard targets).
\item For DGR+distill, also a separate generative model was trained to generate the images to be replayed, but these were then paired with soft targets (as in LwF) instead of hard targets (as in DGR).
\end{itemize}
\item[- iCaRL:] This method was implemented following \citet{rebuffi2017icarl}; see section~\ref{sec:icarl} in the Appendix for details. For the results in Tables~\ref{tab:splitMNIST} and \ref{tab:permMNIST}, a memory budget of 2000 was used. Due to the way iCaRL is set up with distillation of current task data on the classes of all previous tasks using binary classification loss, it can only be applied in the Class-IL scenario. However, two components of iCaRL---the use of exemplars for classification and the replay of stored data during training---are suitable for all scenarios. Both these components are explored in section~\ref{sec:exact_replay} in the Appendix.
\end{description}

We included the following two baselines:
\begin{description}
\item[- None:] The model was sequentially trained on all tasks in the standard way. This is also called \emph{fine-tuning}, and can be seen as a lower bound.
\item[- Offline:] The model was always trained using the data of all tasks so far. This is also called \emph{joint training}, and was included as it can be seen as an upper bound.
\end{description}

All methods except for iCaRL used the standard multi-class cross entropy classification loss for the model's predictions on the current task data ($\mathcal{L}_{\text{current}} = \mathcal{L}_{\text{classification}}$). For the split MNIST protocol, all models were trained for 2000 iterations per task using the ADAM-optimizer ($\beta_1=0.9$, $\beta_2=0.999$; \citep{kingma2014adam}) with learning rate 0.001. The same optimizer was used for the permuted MNIST protocol, but with 5000 iterations and learning rate 0.0001. For each iteration, $\mathcal{L}_{\text{current}}$ (and $\mathcal{L}_{\text{regularization}}$) was calculated as average over 128 samples from the current task. If replay was used, in each iteration also 128 replayed samples were used to calculate $\mathcal{L}_{\text{replay}}$.

For DGR and DGR+distill, a separate generative model was sequentially trained on all tasks. A symmetric variational autoencoder [VAE; \citealp{kingma2013auto}] was used as generative model, with 2 fully connected hidden layers of 400 (split MNIST) or 1000 (permuted MNIST) units and a stochastic latent variable layer of size 100. A standard normal distribution was used as prior. See section \ref{sec:VAE} in the Appendix for more details. Training of the generative model was also done with generative replay (provided by its own copy stored after finishing training on the previous task) and with the same hyperparameters (i.e., learning rate, optimizer, iterations, batch sizes) as for the main model.

\section{Results}
\label{sec:results}

For the split MNIST task protocol, we found a clear difference in difficulty between the three continual learning scenarios (Table \ref{tab:splitMNIST}). All of the tested methods performed well in the Task-IL scenario, but LwF and especially the regularization-based methods (EWC, Online EWC and SI) struggled in the Domain-IL scenario and completely failed in the Class-IL scenario. Importantly, only methods using replay (DGR, DGR+distill and iCaRL) obtained good performance (above 90\%) in the Domain-IL and Class-IL scenarios. Somewhat strikingly, we found that in all scenarios, replaying images from the current task (LwF; e.g., replaying `2's and `3's in order not to forget how to recognize `0's and `1's), prevented the forgetting of previous tasks better than any of the regularization-based methods. We further note that in contrast to several recent reports [e.g., \citealp{nguyen2017variational,van2018generative,hsu2018re}], we obtained competitive performance of EWC (and Online EWC) on the task-IL scenario variant of split MNIST (i.e., `multi-headed split MNIST'). Reason for this difference is that we explored a much wider hyperparameter range; our selected values were several orders of magnitude larger than those typically considered (see section~\ref{sec:hyper} in the Appendix).\footnote{An explanation for these extreme hyperparameter values is that the individual tasks of the split MNIST protocol (i.e., distinguishing between two digits) are relatively easy, making that after finishing training on each task the gradients---and thus the Fisher Information on which EWC is based---are very small.}

\begin{table*}[t]
  \caption{\label{tab:splitMNIST}Average test accuracy (over all tasks) on the split MNIST task protocol. Each experiment was performed 20 times with different random seeds, reported is the mean ($\pm$ SEM) over these runs.}
  \begin{center}
  \begin{small}
  \begin{tabular}{llp{2.11cm}p{2.11cm}p{2.11cm}} \toprule
    \textbf{Approach} & \textbf{Method} & \textbf{Task-IL} & \textbf{Domain-IL} & \textbf{Class-IL} \\ \midrule \midrule
    \multirow{2}{*}{\it Baselines} & \it None  --  lower bound & \it 87.19 ($\pm$ 0.94) & \it 59.21 ($\pm$ 2.04) & \it 19.90 ($\pm$ 0.02) \\
    & \it Offline  --  upper bound & \it 99.66 ($\pm$ 0.02) & \it 98.42 ($\pm$ 0.06) & \it 97.94 ($\pm$ 0.03) \\ \cmidrule[0.7pt]{1-5}
    Task-specific & XdG & 99.10 ($\pm$ 0.08) & - & - \\ \midrule
    \multirow{3}{*}{Regularization} & EWC & 98.64 ($\pm$ 0.22) & 63.95 ($\pm$ 1.90) & 20.01 ($\pm$ 0.06) \\
    & Online EWC & 99.12 ($\pm$ 0.11) & 64.32 ($\pm$ 1.90) & 19.96 ($\pm$ 0.07) \\
    & SI & 99.09 ($\pm$ 0.15) & 65.36 ($\pm$ 1.57) & 19.99 ($\pm$ 0.06) \\
    \midrule
    \multirow{3}{*}{Replay} & LwF & 99.57 ($\pm$ 0.02) & 71.50 ($\pm$ 1.63) & 23.85 ($\pm$ 0.44) \\
    & DGR & 99.50 ($\pm$ 0.03) & 95.72 ($\pm$ 0.25) & 90.79 ($\pm$ 0.41) \\
    & DGR+distill & 99.61 ($\pm$ 0.02) & 96.83 ($\pm$ 0.20) & 91.79 ($\pm$ 0.32) \\ \midrule
    Replay + Exemplars & iCaRL (budget = 2000) & - & - & 94.57 ($\pm$ 0.11) \\
    \bottomrule
  \end{tabular}
  \end{small}
  \end{center}
  \vskip -0.1in
\end{table*}

\begin{table*}[t]
  \caption{\label{tab:permMNIST}Idem as Table \ref{tab:splitMNIST}, except on the permuted MNIST task protocol.}
  \begin{center}
  \begin{small}
  \begin{tabular}{llp{2.11cm}p{2.11cm}p{2.11cm}} \toprule
    \textbf{Approach} & \textbf{Method} & \textbf{Task-IL} & \textbf{Domain-IL} & \textbf{Class-IL} \\ \midrule \midrule
    \multirow{2}{*}{\it Baselines} & \it None -- lower bound & \it 81.79 ($\pm$ 0.48) & \it 78.51 ($\pm$ 0.24) & \it 17.26 ($\pm$ 0.19) \\
    & \it Offline -- upper bound & 97.68 ($\pm$ 0.01) & 97.59 ($\pm$ 0.01) & 97.59 ($\pm$ 0.02) \\ \cmidrule[0.7pt]{1-5}
    Task-specific & XdG & 91.40 ($\pm$ 0.23) & - & - \\ \midrule
    \multirow{3}{*}{Regularization} & EWC & 94.74 ($\pm$ 0.05) & 94.31 ($\pm$ 0.11) & 25.04 ($\pm$ 0.50) \\
    & Online EWC & 95.96 ($\pm$ 0.06) & 94.42 ($\pm$ 0.13) & 33.88 ($\pm$ 0.49) \\
    & SI & 94.75 ($\pm$ 0.14) & 95.33 ($\pm$ 0.11) & 29.31 ($\pm$ 0.62) \\
    \midrule
    \multirow{3}{*}{Replay} & LwF & 69.84 ($\pm$ 0.46) & 72.64 ($\pm$ 0.52) & 22.64 ($\pm$ 0.23) \\
    & DGR & 92.52 ($\pm$ 0.08) & 95.09 ($\pm$ 0.04) & 92.19 ($\pm$ 0.09) \\
    & DGR+distill & 97.51 ($\pm$ 0.01) & 97.35 ($\pm$ 0.02) & 96.38 ($\pm$ 0.03) \\ \midrule
    Replay + Exemplars & iCaRL (budget = 2000) & - & - & 94.85 ($\pm$ 0.03) \\
    \bottomrule
  \end{tabular}
  \end{small}
  \end{center}
  \vskip -0.1in
\end{table*}

For the permuted MNIST protocol, all methods except LwF performed well in both the Task-IL and the Domain-IL scenario (Table~\ref{tab:permMNIST}). In the Class-IL, however, the regularization-based methods failed again and only replay-based methods obtained good performance. The difference between the Task-IL and Domain-IL scenarios was only small for this task protocol, but this might be because---except in XdG---task identity information was only used in the output layer, while information about the applied permutation would likely be more useful in the network's lower layers. Confirming this hypothesis, we found that each method's performance in the Task-IL scenario could be improved by combining it with XdG (i.e., using task-identity in the hidden layers; see section~\ref{sec:task_id} in the Appendix). Finally, while LwF had some success with the split MNIST protocol, this method did not work with the permuted MNIST protocol, presumably because now the inputs of the different tasks were uncorrelated due to the random permutations.

\section{Discussion}
\label{sec:discussion}

Catastrophic forgetting is a major obstacle to the development of artificial intelligence applications capable of true lifelong learning \citep{kumaran2016learning,parisi2018continual}, and enabling neural networks to sequentially learn multiple tasks has become a topic of intense research. Yet, despite its scope, this research field is relatively unstructured: even though the same datasets tend to be used, direct comparisons between published methods are difficult. We demonstrate that an important difference between currently used experimental protocols is whether task identity is provided and---if it is not---whether it must be inferred. These two distinctions led us to identify three scenarios for continual learning of varying difficulty. It is hoped that these scenarios will help to provide structure to the continual learning field and make comparisons between studies easier.

For each scenario, we performed a comprehensive comparison of recently proposed methods. An important conclusion is that for the class-incremental learning scenario (i.e., when task identity must be inferred), currently only replay-based methods are capable of producing acceptable results. In this scenario, even for relatively simple task protocols involving the classification of MNIST-digits, regularization-based methods such as EWC and SI completely fail. On the split MNIST task protocol, regularization-based methods also struggle in the domain-incremental learning scenario (i.e., when task identity does not need to be inferred but is also not provided). These results highlight that for the more challenging, ethological-relevant scenarios where task identity is not provided, replay might be an unavoidable tool.

It should be stressed that a limitation of the current study is that MNIST-images are relatively easy to generate. It therefore remains an open question whether generative replay will still be so successful for task protocols with more complicated input distributions. However, promising for generative replay is that the capabilities of generative models are rapidly improving [e.g., \citealp{goodfellow2014generative,oord2016pixel,rezende2015variational}]. Moreover, the good performance of LwF (i.e., replaying inputs from the current task) on the split MNIST task protocol suggests that even if the quality of replayed samples is not perfect, they could still be very helpful. For a further discussion of the scalability of generative replay we refer to \cite{van2018generative}. Finally, as illustrated by iCaRL, an alternative / complement to replaying generated samples could be to store examples from previous tasks and replay those (see section~\ref{sec:exact_replay} in the Appendix for a further discussion).

\subsubsection*{Acknowledgments}
We thank Mengye Ren, Zhe Li and anonymous reviewers for comments on various parts of this work. This research project has been supported by an IBRO-ISN Research Fellowship, by the Lifelong Learning Machines (L2M) program of the Defence Advanced Research Projects Agency (DARPA) via contract number HR0011-18-2-0025 and by the Intelligence Advanced Research Projects Activity (IARPA) via Department of Interior/Interior Business Center (DoI/IBC) contract number D16PC00003. The U.S. Government is authorized to reproduce and distribute reprints for Governmental purposes notwithstanding any copyright annotation thereon. Disclaimer: The views and conclusions contained herein are those of the authors and should not be interpreted as necessarily representing the official policies or endorsements, either expressed or implied, of DARPA, IARPA, DoI/IBC, or the U.S. Government.

\bibliography{references}
\bibliographystyle{unsrtnat}

\newpage

\appendix
\renewcommand\thefigure{\thesection.\arabic{figure}}   
\renewcommand\thetable{\thesection.\arabic{table}}   

\section{Additional experimental details}
\setcounter{figure}{0}  
\setcounter{table}{0}  

PyTorch-code to perform the experiments described in this report is available online: \url{https://github.com/GMvandeVen/continual-learning}.

\subsection{Loss functions}

\subsubsection{Classification}
\label{sec:classifcation}
The standard per-sample cross entropy loss function for an input $\boldsymbol{x}$ labeled with a hard target $y$ is given by:
\begin{equation}
\mathcal{L}_{\text{classification}} \left(\boldsymbol{x},y;\boldsymbol{\theta}\right) = -\log p_{\boldsymbol{\theta}}\left(Y=y|\boldsymbol{x}\right)
\label{eq:class_loss}
\end{equation}
where $p_{\boldsymbol{\theta}}$ is the conditional probability distribution defined by the neural network whose trainable bias and weight parameters are collected in $\boldsymbol{\theta}$. It is important to note that in this report this probability distribution is not always defined over all output nodes of the network, but only over the ``active nodes''. This means that the normalization performed by the final softmax layer only takes into account these active nodes, and that learning is thus restricted to those nodes. For experiments performed according to the Task-IL scenario, for which we use a ``multi-headed'' softmax layer, always only the nodes of the task under consideration are active. Typically this is the current task, but for replayed data it is the task that is (intended to be) replayed. For the Domain-IL scenario always all nodes are active. For the Class-IL scenario, the nodes of all tasks seen so far are active, both when training on current and on replayed data.

For the method DGR, there are also some subtle differences between the continual learning scenarios when generating hard targets for the inputs to be replayed. With the Task-IL scenario, only the classes of the task that is intended to be replayed can be predicted (in each iteration the available replays are equally divided over the previous tasks). With the Domain-IL scenario always all classes can be predicted. With the Class-IL scenario only classes from up to the previous task can be predicted.

\subsubsection{Distillation}
\label{sec:distillation}
The methods LwF and DGR+distill use distillation loss for their replayed data. For this, each input $\boldsymbol{x}$ to be replayed is labeled with a ``soft target'', which is a vector containing a probability for each active class. This target probability vector is obtained using a copy of the main model stored after finishing training on the most recent task, and the training objective is to match the probabilities predicted by the model being trained to these target probabilities (by minimizing the cross entropy between them). Moreover, as is common for distillation, these two probability distributions that we want to match are made softer by temporary raising the temperature $T$ of their models' softmax layers.\footnote{The same temperature should be used for calculating the target probabilities and for calculating the probabilities to be matched during training; but during testing the temperature should be set back to 1. A typical value for this temperature is 2, which is the value used in this report.} This means that before the softmax normalization is performed on the logits, these logits are first divided by $T$. For an input $\boldsymbol{x}$ to be replayed during training of task $K$, the soft targets are given by the vector $\tilde{\boldsymbol{y}}$ whose $c^{\text{th}}$ element is given by:
\begin{equation}
\tilde{y}_c = p_{\hat{\boldsymbol{\theta}}^{(K-1)}}^T\left(Y=c|\boldsymbol{x}\right)
\end{equation}
where $\hat{\boldsymbol{\theta}}^{(K-1)}$ is the vector with parameter values at the end of training of task $K-1$ and $p_{\boldsymbol{\theta}}^T$ is the conditional probability distribution defined by the neural network with parameters $\boldsymbol{\theta}$ and with the temperature of its softmax layer raised to $T$. The distillation loss function for an input $\boldsymbol{x}$ labeled with a soft target vector $\tilde{\boldsymbol{y}}$ is then given by:
\begin{equation}
\mathcal{L}_{\text{distillation}} \left(\boldsymbol{x},\tilde{\boldsymbol{y}};\boldsymbol{\theta}\right) = -T^2 \sum_{c=1}^{N_{\text{classes}}} \tilde{y}_c \log p_{\boldsymbol{\theta}}^T\left(Y=c|\boldsymbol{x}\right)
\label{eq:dist_loss}
\end{equation}
where the scaling by $T^2$ is included to ensure that the relative contribution of this objective matches that of a comparable objective with hard targets \cite{hinton2015distilling}.

When generating soft targets for the inputs to be replayed, there are again subtle differences between the three continual learning scenarios. With the Task-IL scenario, the soft target probability distribution is defined only over the classes of the task intended to be replayed. With the Domain-IL scenario this distribution is always over all classes. With the Clase-IL scenario, the soft target probability distribution is first generated only over the classes from up to the previous task and then zero probabilities are added for all classes in the current task.

\subsection{Regularization terms}
\label{sec:regularization}
\subsubsection{EWC}
\label{sec:EWC}
The regularization term of elastic weight consolidation [EWC; \citealp{kirkpatrick2017overcoming}] consists of a quadratic penalty term for each previously learned task, whereby each task's term penalizes the parameters for how different they are compared to their value directly after finishing training on that task. The strength of each parameter's penalty depends for every task on how important that parameter was estimated to be for that task, with higher penalties for more important parameters. For EWC, a parameter's importance is estimated for each task by the parameter's corresponding diagonal element of that task's Fisher Information matrix, evaluated at the optimal parameter values after finishing training on that task. The EWC regularization term for task $K>1$ is given by:
\begin{equation}
  \mathcal{L}^{(K)}_{\text{regularization}_{\text{EWC}}}\left(\boldsymbol{\theta}\right) = \sum_{k=1}^{K-1} \left( \frac{1}{2} \sum_{i=1}^{N_{\text{params}}} F_{ii}^{(k)} \left(\theta_i - \hat{\theta}_{i}^{(k)} \right)^2 \right)
  \label{eq:ewc}
\end{equation}
whereby $\hat{\theta}_{i}^{(k)}$ is the $i^{\text{th}}$ element of $\hat{\boldsymbol{\theta}}^{\left(k\right)}$, which is the vector with parameter values at the end of training of task $k$, and $F_{ii}^{(k)}$ is the $i^{\text{th}}$ diagonal element of $\boldsymbol{F}^{(k)}$, which is the Fisher Information matrix of task $k$ evaluated at $\hat{\boldsymbol{\theta}}^{(k)}$. Following the definitions and notation in \citet{martens2014new}, the $i^{\text{th}}$ diagonal element of $\boldsymbol{F}^{(k)}$ is defined as:
\begin{equation}
  F_{ii}^{(k)} = \left. \mathbb{E}_{\boldsymbol{x}\sim Q_{\boldsymbol{x}}^{(k)}} \ \mathbb{E}_{p_{\boldsymbol{\theta}}\left(y|\boldsymbol{x}\right)} \left[ \left( \frac{\delta \log{p_{\boldsymbol{\theta}}\left(Y=y|\boldsymbol{x}\right)}}{\delta \theta_i} \right)^2 \right] \right\rvert_{\boldsymbol{\theta}=\hat{\boldsymbol{\theta}}^{(k)}}
  \label{eq:fi_theory}
\end{equation}
whereby $Q_{\boldsymbol{x}}^{(k)}$ is the (theoretical) input distribution of task $k$ and $p_{\boldsymbol{\theta}}$ is the conditional distribution defined by the neural network with parameters $\boldsymbol{\theta}$. Note that in \citet{kirkpatrick2017overcoming} it is not specified exactly how these $F_{ii}^{(k)}$ are calculated (except that it is said to be ``easy''); but we have been made aware that they are calculated as the diagonal elements of the ``true Fisher Information'':
\begin{equation}
  F_{ii}^{(k)} = \frac{1}{|S^{(k)}|} \sum_{\boldsymbol{x}\in S^{(k)}} \left( \left. \frac{\delta\log p_{\boldsymbol{\theta}}\left(Y=\hat{y}_{\boldsymbol{x}}^{(k)}|\boldsymbol{x}\right)}{\delta\theta_i} \right\rvert_{\boldsymbol{\theta}=\hat{\boldsymbol{\theta}}^{(k)}} \right)^2
  \label{eq:emp_fi}
\end{equation}
whereby $S^{(k)}$ is the training data of task $k$ and $\hat{y}_{\boldsymbol{x}}^{(k)} = \mathop{\argmin}_{y} \log p_{\hat{\boldsymbol{\theta}}^{(k)}}\left(Y=y|\boldsymbol{x}\right)$, the label predicted by the model with parameters $\hat{\boldsymbol{\theta}}^{(k)}$ given $\boldsymbol{x}$.\footnote{An alternative way to calculate $F_{ii}^{(k)}$ would be, instead of taking for each training input $\boldsymbol{x}$ only the most likely label predicted by model $p_{\hat{\boldsymbol{\theta}}^{(k)}}$, to sample for each $\boldsymbol{x}$ multiple labels from the entire conditional distribution defined by this model (i.e., to approximate the inner expectation of equation \ref{eq:fi_theory} for each training sample $\boldsymbol{x}$ with Monte Carlo sampling from $p_{\hat{\boldsymbol{\theta}}^{(k)}}\left(\cdot|\boldsymbol{x}\right)$). Another option is to use the ``empirical Fisher Information", by replacing in equation \ref{eq:emp_fi} the predicted label $\hat{y}_{\boldsymbol{x}}^{(k)}$ by the observed label $y$. The results reported in Tables \ref{tab:splitMNIST} and \ref{tab:permMNIST} do not depend much on the choice of how $F_{ii}^{(k)}$ is calculated.} The calculation of the Fisher Information is time-consuming, especially if tasks have a lot of training data. In practice it might therefore sometimes be beneficial to trade accuracy for speed by using only a subset of a task's training data for this calculation (e.g., by introducing another hyperparameter $N_{\text{Fisher}}$ that sets the maximum number of samples to be used in equation~\ref{eq:emp_fi}).

\subsubsection{Online EWC}
\label{sec:online_ewc}
A disadvantage of the original formulation of EWC is that the number of quadratic terms in its regularization term grows linearly with the number of tasks. This is an important limitation, as for a method to be applicable in a true lifelong learning setting its computational cost should not increase with the number of tasks seen so far. It was pointed out by \citet{huszar2018note} that a slightly stricter adherence to the approximate Bayesian treatment of continual learning, which had been used as motivation for EWC, actually results in only a single quadratic penalty term on the parameters that is anchored at the optimal parameters after the most recent task and with the weight of the parameters' penalties determined by a running sum of the previous tasks' Fisher Information matrices. This insight was adopted by \citet{schwarz2018progress}, who proposed a modification to EWC called \emph{online EWC}. The regularization term of online EWC when training on task $K>1$ is given by:
\begin{equation}
  \mathcal{L}^{(K)}_{\text{regularization}_{\text{oEWC}}} = \sum_{i=1}^{N_{\text{params}}} \tilde{F}_{ii}^{(K-1)} \left(\theta_i - \hat{\theta}_{i}^{(K-1)} \right)^2
  \label{eq:online_ewc}
\end{equation}
whereby $\hat{\theta}_i^{(K-1)}$ is the value of parameter $i$ after finishing training on task $K-1$ and $\tilde{F}_{ii}^{(K-1)}$ is a running sum of the $i^{\text{th}}$ diagonal elements of the Fisher Information matrices of the first $K-1$ tasks, with a hyperparameter $\gamma\leq 1$ that governs a gradual decay of each previous task's contribution. That is: $\tilde{F}_{ii}^{(k)} = \gamma \tilde{F}_{ii}^{(k-1)} + F_{ii}^{(k)}$, with $\tilde{F}_{ii}^{(1)} = F_{ii}^{(1)}$ and $F_{ii}^{(k)}$ is the $i^{\text{th}}$ diagonal element of the Fisher Information matrix of task $k$ calculated according to equation \ref{eq:emp_fi}.

\subsubsection{SI}
\label{sec:SI}
Similar as for online EWC, the regularization term of synaptic intelligence [SI; \citealp{zenke2017improved}] consists of only one quadratic term that penalizes changes to parameters away from their values after finishing training on the previous task, with the strength of each parameter's penalty depending on how important that parameter is thought to be for the tasks learned so far. To estimate parameters' importance, for every new task $k$ a per-parameter contribution to the change of the loss is first calculated for each parameter $i$ as follows:
\begin{equation}
  \omega_i^{(k)} = \sum_{t=1}^{N_{\text{iters}}} \left(\theta_i[t^{(k)}]-\theta_i[\left(t-1\right)^{(k)}]\right) \frac{-\delta\mathcal{L}_{\text{total}}[t^{(k)}]}{\delta\theta_i}
  \label{eq:small_omega}
\end{equation}
with $N_{\text{iters}}$ the total number of iterations per task, $\theta_i[t^{(k)}]$ the value of the $i^{\text{th}}$ parameter after the $t^{\text{th}}$ training iteration on task $k$ and $\frac{\delta\mathcal{L}_{\text{total}}[t^{(k)}]}{\delta\theta_i}$ the gradient of the loss with respect to the $i^{\text{th}}$ parameter during the $t^{\text{th}}$ training iteration on task $k$. For every task, these per-parameter contributions are normalized by the square of the total change of that parameter during training on that task plus a small dampening term $\xi$ (set to 0.1, to bound the resulting normalized contributions when a parameter's total change goes to zero), after which they are summed over all tasks so far. The estimated importance of parameter $i$ for the first $K-1$ tasks is thus given by:
\begin{equation}
  \Omega_{i}^{(K-1)} = \sum_{k=1}^{K-1} \frac{\omega_i^{(k)}}{\left(\Delta_i^{(k)}\right)^2+\xi}
  \label{eq:omega}
\end{equation}
with $\Delta_i^{(k)} = \theta_i[{N_{\text{iters}}}^{(k)}]-\theta_i[0^{(k)}]$, where $\theta_i[0^{(k)}]$ indicates the value of parameter $i$ right before starting training on task $k$. (An alternative formulation is $\Delta_i^{(k)} = \hat{\theta}_i^{(k)}-\hat{\theta}_i^{(k-1)}$, with $\hat{\theta}_i^{(0)}$ the value of parameter $i$ it was initialized with and $\hat{\theta}_i^{(k)}$ its value after finishing training on task $k$.) The regularization term of SI to be used during training on task $K$ is then given by:
\begin{equation}
  \mathcal{L}^{(K)}_{\text{regularization}_{\text{SI}}} = \sum_{i=1}^{N_{\text{params}}} \Omega_{i}^{(K-1)} \left(\theta_i - \hat{\theta}_{i}^{(K-1)} \right)^2
  \label{eq:si}
\end{equation}

\subsection{Generative Model}
\label{sec:VAE}
The separate generative model that is used for DGR and DGR+distill is a variational autoencoder (VAE; \citealp{kingma2013auto}), of which both the encoder network $q_{\boldsymbol{\phi}}$ and the decoder network $p_{\boldsymbol{\psi}}$ are multi-layer perceptrons with 2 hidden layers containing 400 (split MNIST) or 1000 (permuted MNIST) units with ReLU non-linearity. The stochastic latent variable layer $\boldsymbol{z}$ has 100 units and the prior over them is the standard normal distribution. Following \citet{kingma2013auto}, the ``latent variable regularization term'' of this VAE is given by:
\begin{equation}
  \mathcal{L}_{\text{latent}}(\boldsymbol{x};\boldsymbol{\phi}) = \frac{1}{2}\sum_{j=1}^{100}\left(1+\log\left({\sigma_j^{(\boldsymbol{x})}}^2\right)-{\mu_j^{(\boldsymbol{x})}}^2-{\sigma_j^{(\boldsymbol{x})}}^2\right)
  \label{eq:vae_latent}
\end{equation}
whereby $\mu_j^{(\boldsymbol{x})}$ and $\sigma_j^{(\boldsymbol{x})}$ are the $j^{\text{th}}$ elements of respectively $\boldsymbol{\mu}^{(\boldsymbol{x})}$ and $\boldsymbol{\sigma}^{(\boldsymbol{x})}$, which are the outputs of the encoder network $q_{\boldsymbol{\phi}}$ given input $\boldsymbol{x}$. Following \citet{doersch2016tutorial}, the output layer of the decoder network $p_{\boldsymbol{\psi}}$ has a sigmoid non-linearity and the ``reconstruction term'' is given by the binary cross entropy between the original and decoded pixel values:
\begin{equation}
  \mathcal{L}_{\text{recon}}\left(\boldsymbol{x};\boldsymbol{\phi},\boldsymbol{\psi}\right) = \sum_{p=1}^{N_{\text{pixels}}} x_p \log\left(\tilde{x}_p\right) + \left(1-x_p\right) \log\left(1-\tilde{x}_p\right)
  \label{eq:vae_recon}
\end{equation}
whereby $x_p$ is the value of the $p^{\text{th}}$ pixel of the original input image $\boldsymbol{x}$ and $\tilde{x}_p$ is the value of the $p^{\text{th}}$ pixel of the decoded image $\tilde{\boldsymbol{x}} = p_{\boldsymbol{\psi}}\left(\boldsymbol{z}^{(\boldsymbol{x})}\right)$ with $\boldsymbol{z}^{(\boldsymbol{x})} = \boldsymbol{\mu}^{(\boldsymbol{x})} + \boldsymbol{\sigma}^{(\boldsymbol{x})} \cdot \boldsymbol{\epsilon}$, whereby $\boldsymbol{\epsilon}$ is sampled from $\mathcal{N}\left(0,\boldsymbol{I}_{100}\right)$. The per-sample loss function for an input $\boldsymbol{x}$ is then given by \citep{kingma2013auto}:
\begin{equation}
  \mathcal{L}_{\text{generative}} \left(\boldsymbol{x};\boldsymbol{\phi},\boldsymbol{\psi}\right) = \mathcal{L}_{\text{recon}}\left(\boldsymbol{x};\boldsymbol{\phi},\boldsymbol{\psi}\right) + \mathcal{L}_{\text{latent}}\left(\boldsymbol{x};\boldsymbol{\phi}\right)
  \label{eq:vae_loss}
\end{equation}
Similar to the main model, the generative model is trained with replay generated by its own copy stored after finishing training on the previous task.

\subsection{iCaRL}
\label{sec:icarl}

\subsubsection{Feature Extractor}
\paragraph{Network Architecture} For our implementation of iCaRL \citep{rebuffi2017icarl}, we used a feature extractor with the same architecture as the neural network used as classifier with the other methods. The only difference is that the softmax output layer was removed. We denote this feature extractor by $\psi_{\boldsymbol{\phi}}(.)$, with its trainable parameters contained in the vector  $\boldsymbol{\phi}$. These parameters were trained based on binary classification / distillation loss (see below). For this---during training only!---a sigmoid output layer was appended to $\psi_{\boldsymbol{\phi}}$. The resulting extended network outputs for any class $c \in \{1, ..., N_{\text{classes so far}}\}$ a binary probability whether input $\boldsymbol{x}$ belongs to it:
\begin{equation}
 p_{\boldsymbol{\theta}}^{c}(\boldsymbol{x}) = \frac{1}{1+e^{-\boldsymbol{w}_{c}^{T}\psi_{\boldsymbol{\phi}}(\boldsymbol{x})}}
\end{equation}
with $\boldsymbol{\theta}=\left(\boldsymbol{\phi}, \boldsymbol{w}_{1}, ..., \boldsymbol{w}_{N_{\text{classes so far}}}\right)$ a vector containing all iCaRL's trainable parameters. Whenever a new class $c$ was encountered, new parameters $\boldsymbol{w}_c$ were added to $\boldsymbol{\theta}$.

\paragraph{Training}
On each task, the parameters in $\boldsymbol{\theta}$ were trained on an extended dataset containing the current task's training data as well as all stored data from previous tasks (see section~\ref{sec:stored_data}). When training on task $K$, each input $\boldsymbol{x}$ with hard target $y$ in this extended dataset is paired with a new target-vector $\boldsymbol{\bar{y}}$ whose $j^{\text{th}}$ element is given by:
\begin{equation}
 \bar{y}_c = \begin{cases} \mbox{$p_{\hat{\boldsymbol{\theta}}^{(K-1)}}^c\left(\boldsymbol{x}\right)$} & \mbox{if class $c$ in task $1, ..., K-1$} \\ \mbox{$\delta_{y=c}$} & \mbox{if class $c$ in task $K$} \end{cases}
\end{equation}
whereby $\hat{\boldsymbol{\theta}}^{(K-1)}$ is the vector with parameter values at the end of training of task $K-1$. The per-sample loss function for an input $\boldsymbol{x}$ labeled with such an ``old-task-soft-target / new-task-hard-target" vector $\boldsymbol{\bar{y}}$ is then given by:
\begin{equation}
  \mathcal{L}_{\text{iCaRL}} \left(\boldsymbol{x},\boldsymbol{\bar{y}};\boldsymbol{\theta}\right) = -\sum_{c=1}^{N_{\text{classes so far}}} \left[ \bar{y}_c \log p_{\boldsymbol{\theta}}^c(\boldsymbol{x}) + \left(1-\bar{y}_c\right) \log \left(1- p_{\boldsymbol{\theta}}^c(\boldsymbol{x})\right) \right]
  \label{eq:icarl_loss}
\end{equation}

\subsubsection{Selection of Stored Data}
\label{sec:stored_data}
The assumption under which iCaRL operates is that up to $B$ data-points (referred to as `exemplars') are allowed to be stored in memory. The available memory budget is evenly distributed over the classes seen so far, resulting in $m=\left\lfloor\frac{B}{N_{\text{classes so far}}}\right\rfloor$ stored exemplars per class. After training on a task is finished, the selection of data stored in memory is updated as follows:

\paragraph{Create exemplar-sets for new classes} For each new class $c$, iteratively $m$ exemplars are selected based on their extracted feature vectors according to a procedure referred to as `herding'. In each iteration, a new example from class $c$ is selected such that the average feature vector over all selected examples is as close as possible to the average feature vector over all available examples of class $c$. Let $\mathcal{X}^c=\{\boldsymbol{x}_1, ..., \boldsymbol{x}_{N_c}\}$ be the set of all available examples of class $c$ and let $\boldsymbol{\mu}^c=\frac{1}{N_c}\sum_{\boldsymbol{x} \in \mathcal{X}^c}\psi_{\boldsymbol{\phi}}(\boldsymbol{x})$ be the average feature vector over set $\mathcal{X}^c$. The $n^{\text{th}}$ exemplar (for $n=1, ..., m$) to be selected for class $c$ is then given by:
\begin{equation}
 \boldsymbol{p}_{n}^{c} = \argmin_{\boldsymbol{x} \in \mathcal{X}^c} \left\|\boldsymbol{\mu}^c-\frac{1}{n}\left(\psi_{\boldsymbol{\phi}}(\boldsymbol{x})+\sum_{j=1}^{n-1}\psi_{\boldsymbol{\phi}}(\boldsymbol{p}_{j}^{c})\right)\right\|
\end{equation}
This results in ordered exemplar-sets $\mathcal{P}^c = \{\boldsymbol{p}_{1}^{c}, ..., \boldsymbol{p}_{m}^{c}\}$ for each new class $c$.

\paragraph{Reduce exemplar-sets for old classes} If the existing exemplar-sets for the old classes contain more than $m$ exemplars each, for each old class the last selected exemplars are discarded until only $m$ exemplars per class are left.

\subsubsection{Nearest-Class-Mean Classification}
Classification by iCaRL is performed according to a nearest-class-mean rule in feature space based on the stored exemplars. For this, let $\boldsymbol{\mu}_c=\frac{1}{\left|\mathcal{P}^c\right|} \sum_{\boldsymbol{p} \in \mathcal{P}^c} \psi_{\boldsymbol{\phi}}(\boldsymbol{p})$ for $c=1,...,N_{\text{classes so far}}$. The label $y^*$ predicted for a new input $\boldsymbol{x}$ is then given by:
\begin{equation}
 y^* = \argmin_{c=1,...,N_{\text{classes so far}}} \left\|\psi_{\boldsymbol{\phi}}(\boldsymbol{x})-\boldsymbol{\mu}_c\right\|
\end{equation}

\section{Using Task Identity in Hidden Layers}
\label{sec:task_id}
\setcounter{figure}{0}
\setcounter{table}{0}

For the permuted MNIST protocol, there were only small differences in Table~\ref{tab:permMNIST} between the results of the Task-IL and Domain-IL scenario. This suggests that for this protocol, it is actually not so important whether task identity information is available at test time. However, in the main text it was hypothesized that this was the case because for most of the methods---only exception being XdG---task identity was only used in the network's output layer, while information about which permutation was applied is likely more useful in the lower layers. To test this, we performed all methods again on the Task-IL scenario of the permuted MNIST protocol, this time instead using task identity information in the network's hidden layers by combining each method with XdG. This significantly increased the performance of every method (Table~\ref{tab:task_id}), thereby demonstrating that also for the permuted MNIST protocol, the Task-IL scenario is indeed easier than the Domain-IL scenario.

\begin{table}[h]
  \caption{\label{tab:task_id}Comparing two ways of using task identity information in the Task-IL scenario of the permuted MNIST task protocol. In the first column, each method uses a separate output layer for each task (i.e., multi-headed output layer; same as in Table~\ref{tab:permMNIST}). In the second column, each method is instead combined with XdG.}
  \vskip 0.15in
  \begin{center}
  \begin{tabular}{lp{2.2cm}p{2.2cm}} \toprule
    \multirow{2}{*}{\bf Method} & \textbf{\textit{+ task-ID in}} & \textbf{\textit{+ task-ID in}} \\
    & \textbf{\textit{output layer}} & \textbf{\textit{hidden layers}} \\
    \midrule \midrule
    None & 81.79 ($\pm$ 0.48) & 90.41 ($\pm$ 0.32) \\
    \midrule
    EWC & 94.74 ($\pm$ 0.05) & 96.94 ($\pm$ 0.02) \\
    Online EWC & 95.96 ($\pm$ 0.06) & 96.89 ($\pm$ 0.03) \\
    SI & 94.75 ($\pm$ 0.14) & 96.53 ($\pm$ 0.04) \\
    \midrule
    LwF & 69.84 ($\pm$ 0.46) & 84.21 ($\pm$ 0.48) \\
    DGR & 92.52 ($\pm$ 0.08) & 95.31 ($\pm$ 0.04) \\
    DGR+distill & 97.51 ($\pm$ 0.01) & 97.67 ($\pm$ 0.01) \\
    \bottomrule
  \end{tabular}
  \end{center}
  \vskip -0.1in
\end{table}

\section{Replay of Stored Data}
\label{sec:exact_replay}

\begin{figure}[h]
  \vskip 0.15in
  \begin{center}
  \centerline{\includegraphics[width=0.79\columnwidth]{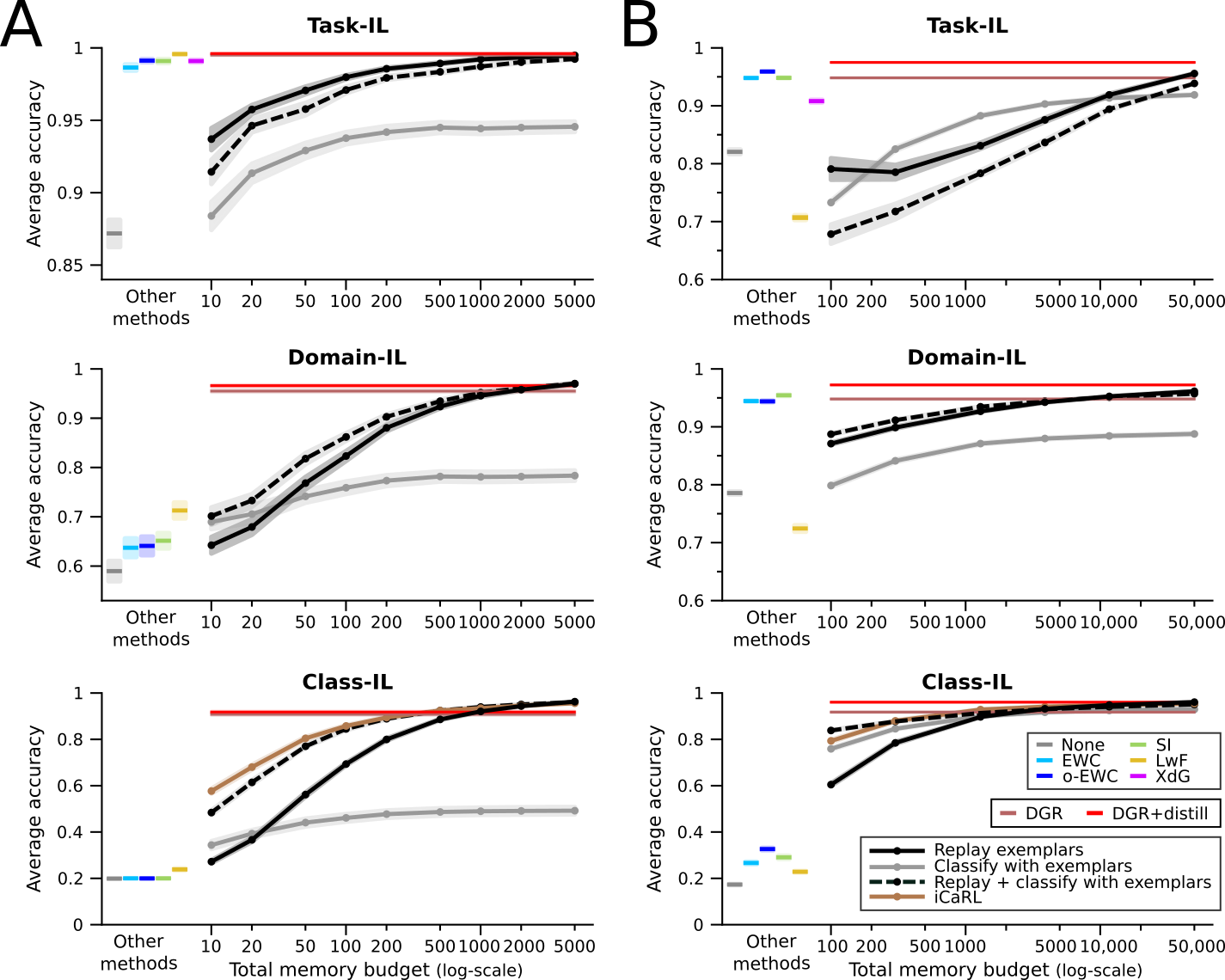}}
  \caption{\label{fig:exact_replay} Average test accuracy (over all tasks) of different ways of using stored data on split MNIST (A) and on permuted MNIST (B) as a function of the total number of examples allowed to be stored in memory. For comparison, the accuracy of the other methods is indicated on the left or as horizontal lines. Displayed are the means over 20 repetitions, shaded areas are $\pm$~1 SEM.}
  \end{center}
  \vskip -0.1in
\end{figure}

As an alternative to generative replay, when data from previous tasks can be stored, it is possible to instead replay that data during training on new tasks. Another way in which stored data could be used is during execution: for example, as in iCaRL, instead of using the softmax classifier, classification can be done using a nearest-class-mean rule in feature space with class means calculated based on the stored data. Stored data can also be used during both training and execution. We evaluated the performance of these different variants of exact replay as a function of the total number of examples allowed to be stored in memory (Figure~\ref{fig:exact_replay}), whereby the data to be stored was always selected using the same `herding'-algorithm as used in iCaRL (see section~\ref{sec:stored_data}).

In the Class-IL scenario, we found that for both task protocols even storing one example per class was enough for any of the exact replay methods to outperform all regularization-based methods. However, to match the performance of generative replay, it was required to store substantially more data. In particular, for permuted MNIST, even with 50,000 stored examples exact replay variants were consistently outperformed by DGR+distill.

\section{Hyperparameters}
\label{sec:hyper}
\setcounter{figure}{0}
\setcounter{table}{0}  

\begin{figure}
  \begin{center}
  \includegraphics[width=0.88\textwidth]{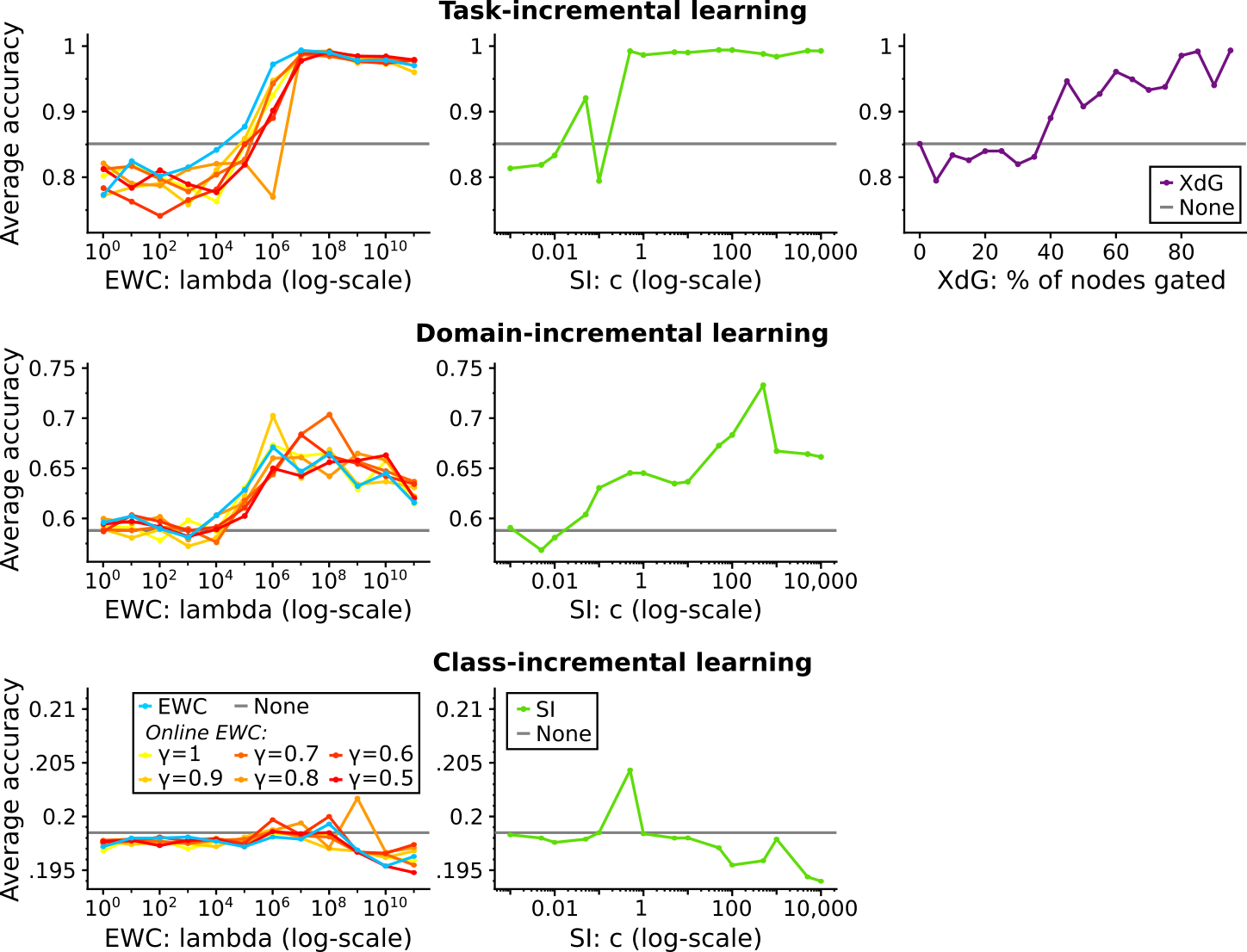}
  \caption{\label{fig:gridSplit}Grid searches for the split MNIST task protocol. Shown are the average test set accuracies (over all 5 tasks) for the (combination of) hyperparameter-values tested for each method.}
  \end{center}
\end{figure}

\begin{figure}
  \begin{center}
  \includegraphics[width=0.88\textwidth]{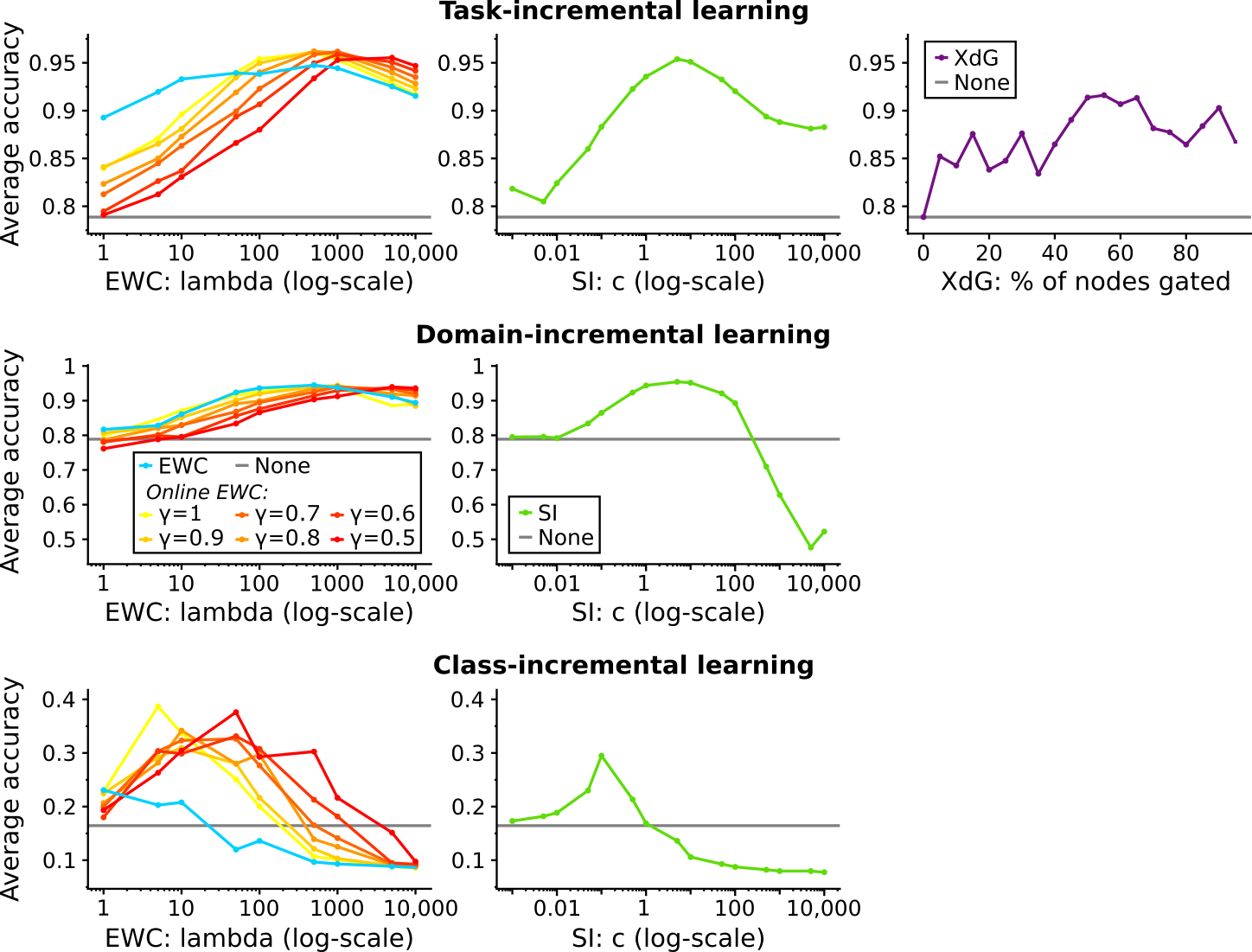}
  \caption{\label{fig:gridPerm}Grid searches for the permuted MNIST task protocol. Shown are the average test set accuracies (over all 10 tasks) for the (combination of) hyperparameter-values tested for each method.}
  \end{center}
\end{figure}

Several of the in this report compared continual learning methods have one or more hyperparameters. The typical way of setting the value of hyperparameters is by training models on the training set for a range of hyperparameter-values, and selecting those that result in the best performance on a separate validation set. This strategy has been adapted to the continual learning setting as training models on the full protocol with different hyperparameter-values using only every task's training data, and comparing their overall performances using separate validation sets (or sometimes the test sets) for each task [e.g., see \citealp{goodfellow2013empirical,kirkpatrick2017overcoming,kemker2017measuring,schwarz2018progress}]. However, here we would like to stress that this means that these hyperparameters are set (or learned) based on an evaluation using data from all tasks, which violates the continual learning principle of only being allowed to visit each task once and in sequence. Although it is tempting to think that it is acceptable to relax this principle for tasks' validation data, we argue here that it is not. A clear example of how using each task's validation data continuously throughout an incremental training protocol can lead to an in our opinion unfair advantage is provided by \citet{wu2018incremental}, in which after finishing training on each task a ``bias-removal parameter'' is set that optimizes performance on the validation sets of all tasks seen so far (see their section 3.3). Although the hyperparameters of the methods compared here are much less influential than those in the above report, we believe that it is important to realize this issue associated with traditional grid searches in a continual learning setting and that at a minimum influential hyperparameters should be avoided in methods for continual learning. 

Nevertheless, to give the competing methods of generative replay the best possible chance---and to explore how influential their hyperparameters are---we do perform grid searches to set the values of their hyperparameters (see Figures \ref{fig:gridSplit} and \ref{fig:gridPerm}). Given the issue discussed above we do not see much value in using validation sets for this, and we evaluate the performances of all hyperparameter(-combination)s using the tasks' test sets. For this grid search each experiment is run once, after which 20 new runs are executed using the selected hyperparameter-values to obtain the results in Tables \ref{tab:splitMNIST} and \ref{tab:permMNIST} in the main text.


\end{document}